\documentclass[10pt,twocolumn,english]{./ieeeconf}
\usepackage[T1]{fontenc}
\usepackage[latin9]{inputenc}
\usepackage{geometry}
\usepackage{array}
\usepackage{refstyle}
\usepackage{multirow}
\usepackage{amstext}
\usepackage{graphicx}

\makeatletter

\AtBeginDocument{\providecommand\figref[1]{\ref{fig:#1}}}
\AtBeginDocument{\providecommand\tabref[1]{\ref{tab:#1}}}
\AtBeginDocument{\providecommand\subsecref[1]{\ref{subsec:#1}}}
\AtBeginDocument{\providecommand\Tabref[1]{\ref{Tab:#1}}}
\AtBeginDocument{\providecommand\Figref[1]{\ref{Fig:#1}}}
\providecommand{\tabularnewline}{\\}
\RS@ifundefined{subsecref}
  {\newref{subsec}{name = \RSsectxt}}
  {}
\RS@ifundefined{thmref}
  {\def\RSthmtxt{theorem~}\newref{thm}{name = \RSthmtxt}}
  {}
\RS@ifundefined{lemref}
  {\def\RSlemtxt{lemma~}\newref{lem}{name = \RSlemtxt}}
  {}

\IfFileExists{colortbl.sty}
 {\usepackage{colortbl}}

\definecolor{grey}{rgb}{0.85,0.85,0.85}
\definecolor{green}{rgb}{0.65,0.85,0.7}
\definecolor{brown}{rgb}{0.81,0.76,0.64}
\definecolor{red}{rgb}{0.96,0.81,0.8}
\definecolor{purple}{rgb}{0.77,0.69,0.85}
\definecolor{yellow}{rgb}{1,0.89,0.69}

\usepackage[caption=false,font=footnotesize]{subfig}
\usepackage[font=footnotesize]{caption}
\usepackage[symbol]{footmisc}

\usepackage{float}
\floatstyle{plaintop}
\restylefloat{table}

\@ifundefined{showcaptionsetup}{}{%
 \PassOptionsToPackage{caption=false}{subfig}}
\usepackage{subfig}
\makeatother

\usepackage{babel}
\begin{document}
\title{\textbf{Lane level context and hidden space characterization}\\
\textbf{ for autonomous driving}}
\author{\thanks{\textit{\emph{This work is carried out within SIVALab, a shared laboratory
between Renault and Heudiasyc (UTC/CNRS), and financed by the CNRS.}}}Corentin Sanchez$^{1}$\thanks{$^{1}$Sorbonne universit�s, Universit� de Technologie de Compi�gne,
CNRS UMR 7253 Heudiasyc, France}, Philippe Xu$^{1}$, Alexandre Armand$^{2}$\thanks{$^{2}$Renault S.A.S, Guyancourt, France},
Philippe Bonnifait$^{1}$}
\maketitle
\begin{abstract}
For an autonomous vehicle, situation understanding is a key capability
towards safe and comfortable decision-making and navigation. Information
is in general provided by multiple sources. Prior information about
the road topology and traffic laws can be given by a High Definition
(HD) map while the perception system provides the description of the
space and of road entities evolving in the vehicle surroundings. In
complex situations such as those encountered in urban areas, the road
user behaviors are governed by strong interactions with the others,
and with the road network. In such situations, reliable situation
understanding is therefore mandatory to avoid inappropriate decisions.
Nevertheless, situation understanding is a complex task that requires
access to a consistent and non-misleading representation of the vehicle
surroundings. This paper proposes a formalism (an interaction lane
grid) which allows to represent, with different levels of abstraction,
the navigable and interacting spaces which must be considered for
safe navigation. A top-down approach is chosen to assess and characterize
the relevant information of the situation. On a high level of abstraction,
the identification of the areas of interest where the vehicle should
pay attention is depicted. On a lower level, it enables to characterize
the spatial information in a unified representation and to infer additional
information in occluded areas by reasoning with dynamic objects.
\end{abstract}

\vspace{-8pt}

\section{Introduction}

\begin{figure}[t]
\centering{}\includegraphics[width=0.8\columnwidth]{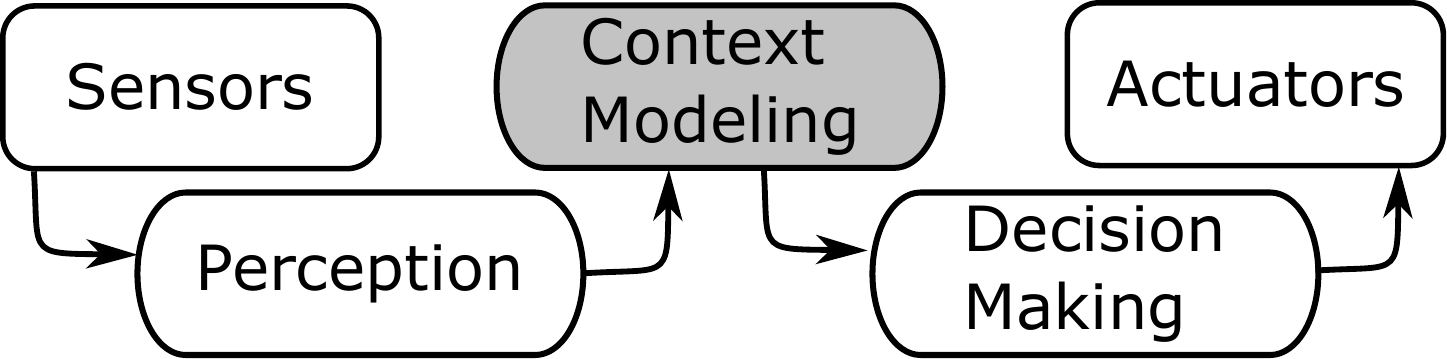}\caption{Place of Context Modeling in a simplified architecture. \label{fig:adv_modules}}
\end{figure}

An autonomous vehicle senses its surrounding environment through its
sensors and continuously takes decisions. Level 3 autonomous driving
systems are able to handle very structured and clear situations like
highways. However, there is still a lack for operating in urban situations
where the decision-making has to be performed in much more complex
situations. Several processes are involved before taking the final
decision. The data provided by perception sensors such as LiDARs or
cameras is processed by perception algorithms in order to supply information
about the surroundings. Then, the situation must be understood so
that the decision-making can plan the maneuver to adopt. The work
presented in this article focuses on the lane level situation modeling
for autonomous driving. In \figref{adv_modules}, the gray module
defines the working area in which this work takes place.

The notions of scene and situation defined in \cite{ulbrichDefiningSubstantiatingTerms2015}
are bases for describing the driving situation context. The selection
process, or ``prioritization'' as described in \cite{refaatAgentPrioritizationAutonomous2019a},
enables to infer the most relevant information. There is a need to
analyze a situation by aggregating all available information from
different sources, \textsl{i.e.}, perception, map, localization. Situation
understanding implies to have a representation which provides non-misleading
information for decision-making. Further, it also allows to reason
by predicting the evolution of the elements of the environment. Depending
on the situation, \textsl{e.g.}, a roundabout or a pedestrian crossing,
the areas of interest where the vehicle has to pay attention are different.
These areas must be identified all along the journey. Relevant road
users that are in interaction or may interact with the vehicle should
be identified while others can be ignored.

A list of agents can be supplied by the perception system as in \cite{hubmannAutomatedDrivingUncertain2018,nohDecisionMakingFrameworkAutonomous2019}
in the form of object maps. However, in case of missed detections,
absence of object does not mean that the space is free as there may
be a lack of visibility. This can lead the vehicle to an inconsistent
situation interpretation. Another approach is to discretize the environment
with grids. Each cell can supply information like occupancy which
can be used by the decision-making to plan the best trajectory \cite{laugierSituationAwarenessDecisionmakinga}.
Evidential grids have the advantage to distinguish a cell that has
not been observed, \textsl{i.e.}, missing information, from a cell
that is conflicting between occupied and non-occupied states \cite{morasMovingObjectsDetection2011}.
However, classification and prediction of dynamic occupied cells is
more complicated, grids are less suited to manage dynamic road users.
The current trend is to use a prior map with several layers like a
geometric layer, a topological layer and a semantic layer (see \cite{poggenhansLanelet2HighdefinitionMap2018})
which can be used jointly with the detected objects as in \cite{ulbrichGraphbasedContextRepresentation2014}
or with the free space characterized by an occupancy grid \cite{mouhagirUsingEvidentialOccupancy2017}.

The environment representation is therefore a key point to understand
the situation and to infer knowledge. Missing information comes from
occlusions \cite{sunBehaviorPlanningAutonomous2019,schornerPredictiveTrajectoryPlanning2019,hubmannPOMDPManeuverPlanner2019,hoermannEnteringCrossroadsBlind2017}
but also from limited range of the vehicle field of view \cite{orzechowskiTacklingOcclusionsLimited2018}.
Depending on the situation encountered and areas of interest, this
will not have the same impact.

This paper aims at defining a lane level grid representation based
on areas of interest. We propose a spatial representation that handles
occlusions and infers additional knowledge in hidden areas to facilitate
safe decision-making. Indeed, some hidden parts do not present any
danger to the ego-vehicle. We therefore propose a formalism that allows
to characterize and manipulate them in real-time. These are our main
contributions. Section \ref{sec:Approach} and \ref{sec:Interaction-objects}
depict the selection of these areas and the classification process.
Finally, some case studies are presented.

\begin{table*}
\begin{tabular}{|c|c|c|c|c|}
\hline 
{\footnotesize{}Interaction type vs} & \multirow{2}{*}{{\footnotesize{}Lane keeping}} & \multirow{2}{*}{{\footnotesize{}Lane Changing}} & \multirow{2}{*}{{\footnotesize{}Lane Merging}} & \multirow{2}{*}{{\footnotesize{}Lane Crossing}}\tabularnewline
{\footnotesize{}Lanes importance} &  &  &  & \tabularnewline
\hline 
{\footnotesize{}Primary order} & \includegraphics[width=50pt,angle=-90]{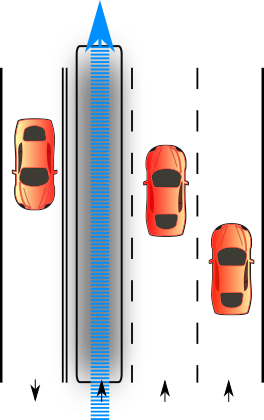} & \includegraphics[angle=-90,width=75pt]{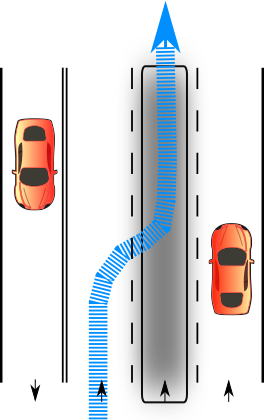} & \includegraphics[angle=-90,width=75pt]{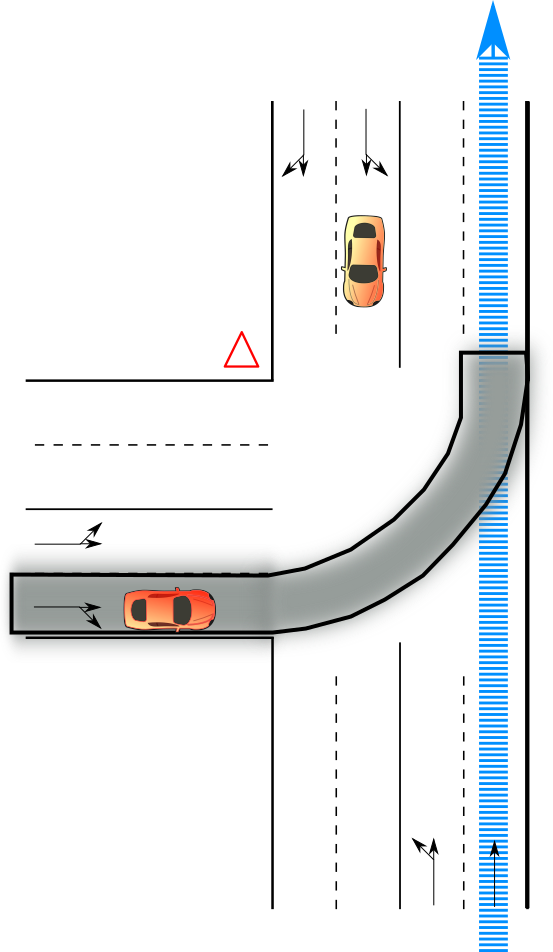} & \includegraphics[angle=-90,width=75pt]{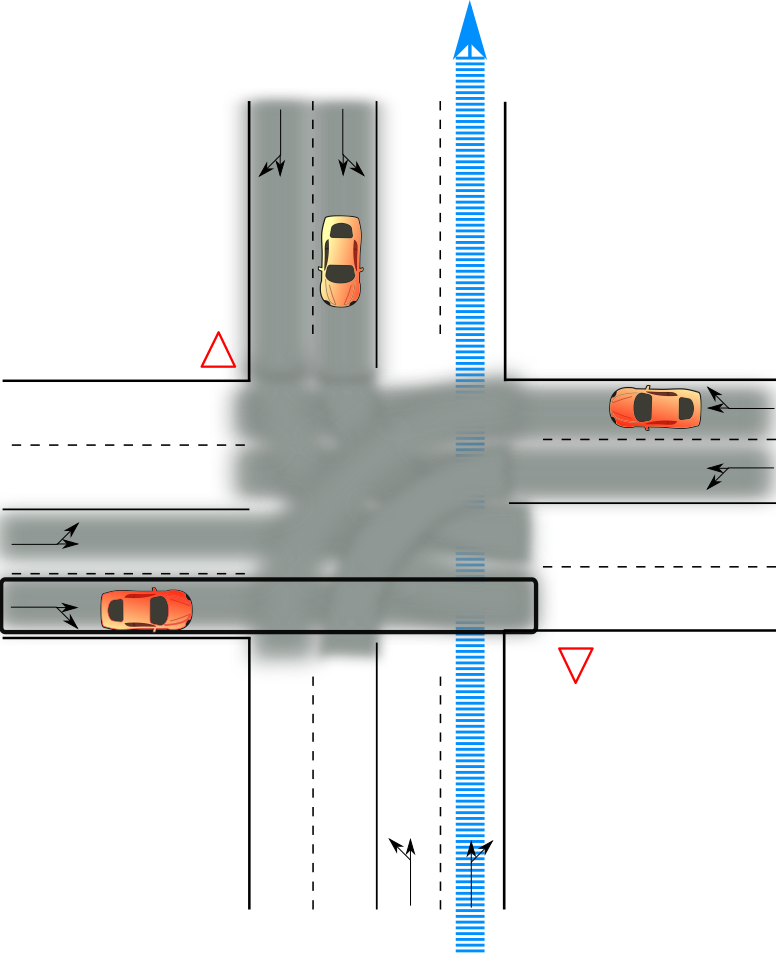}\tabularnewline
\hline 
{\footnotesize{}Secondary order} & \includegraphics[angle=-90,width=75pt]{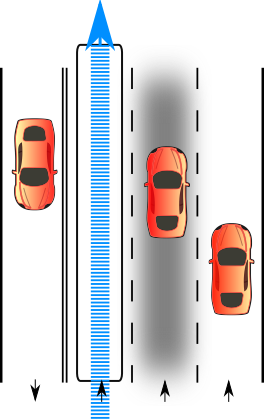} & \includegraphics[angle=-90,width=75pt]{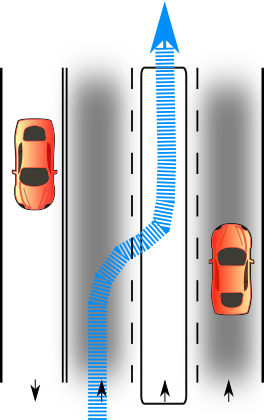} & \includegraphics[angle=-90,width=75pt]{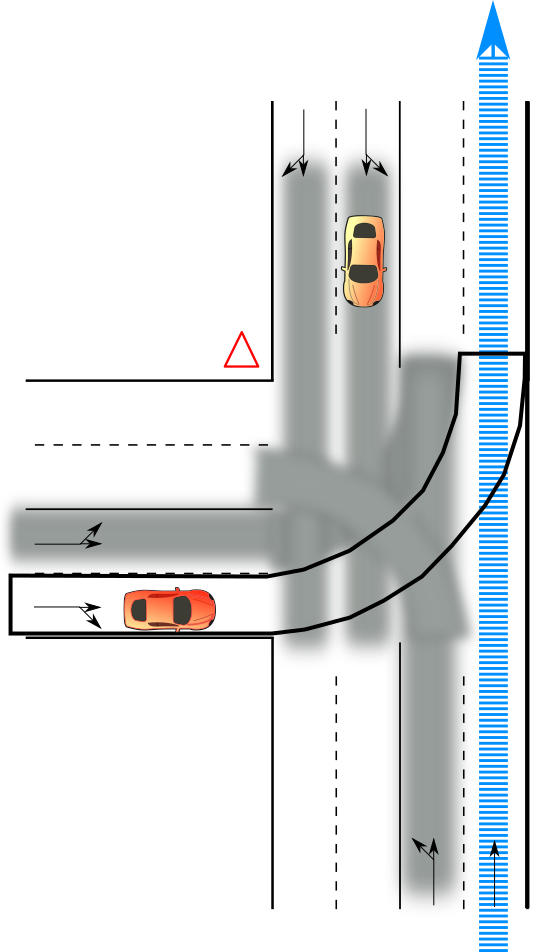} & \includegraphics[angle=-90,width=75pt]{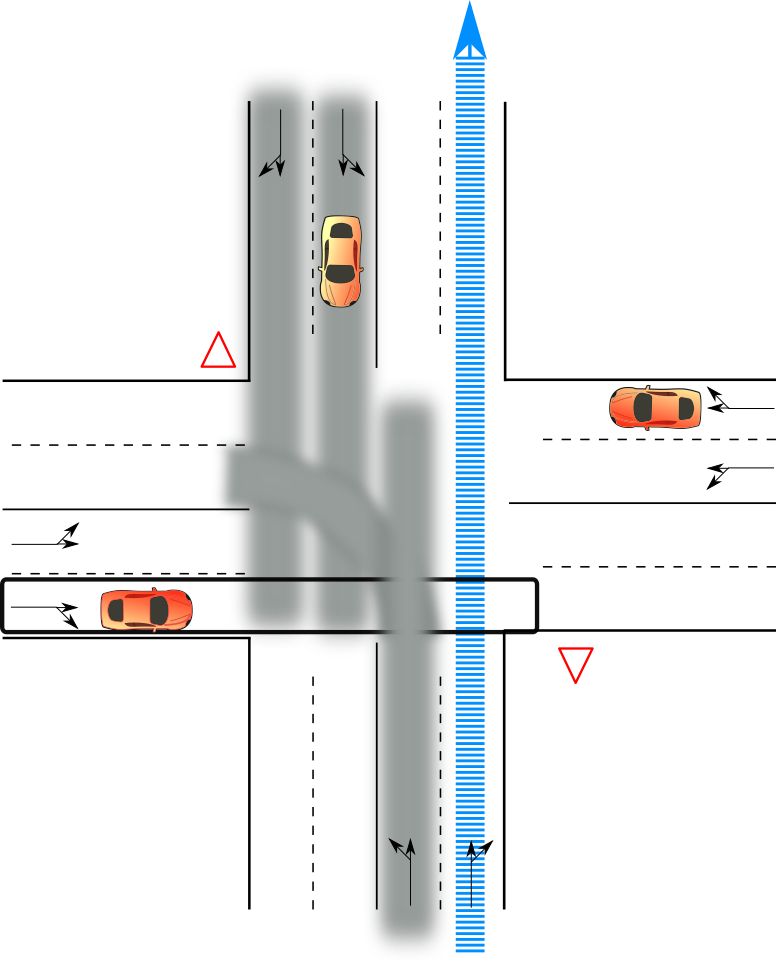}\tabularnewline
\hline 
\end{tabular}
\centering{}\caption{Lanes of interest (road users specific) represented in gray are based
on the longitudinal importance order of the lateral interaction type.
The autonomous vehicle planned path is represented in blue. For visibility
issues the secondary order shows the lanes of interest corresponding
to one lane of the primary order (surrounded in black)\label{tab:Paths-of-interest}}
\end{table*}

\section{Related work}

Decisions operated by the autonomous vehicle are taken at different
levels of abstraction. Early with \cite{michonCriticalViewDriver1985},
the notion of abstraction levels is presented and reused in several
architectures, especially for the decision-making \cite{ulbrichGraphbasedContextRepresentation2014}.
The Operational Level concerns dense information updated at a high
rate and is reactive to events that occur close in time and space,
\textsl{e.g.}, the local trajectory planned with its related speed
profile. The upper level is the tactical level that corresponds to
bigger events which happen in a longer time horizon like maneuvers
\cite{hubmannGenericDrivingStrategy2016}. The most abstracted layer
is the Strategic level. The mission planer plans a global trajectory
over a larger horizon of time and space. 

Each level of abstraction contributes to the representation of the
situation and enables reasoning. In several architectures, all processes
that refer to situation understanding or context modeling notions
are not defined in a unified point of view. In \cite{ulbrichFunctionalSystemArchitecture2017},
the scene modeling and situation modeling are separated whereas in
\cite{tasFunctionalSystemArchitectures2016} they rather stand in
the perception module. The notion of a separated world model module
is used as well \cite{albus4DRCSReference2002,ghetaWorldModelingAutonomous,furdaEnablingSafeAutonomous2011}.
The purpose of such a module is to provide a representation of the
situation of the vehicle, with prediction and analysis to the decision-making.

The representation of information can be done either in a continuous
or discrete manner. In \cite{pappWorldModelingCooperative2008}, the
Local Dynamic Map structures all the objects and features of the world.
A map is also used and as in \cite{poggenhansLanelet2HighdefinitionMap2018},
a metrical layer of information is stored. The notion of a Parametric
Free Space \cite{schreierCompactRepresentationDynamic2016} gives
information on the free space and its contour, \textsl{i.e.}, obstacle
boundary or unknown boundary. When the representation is discrete,
the use of grids is popular \cite{mouhagirUsingEvidentialOccupancy2017,hoermannEnteringCrossroadsBlind2017}.

As presented before, there is a need to combine these representations,
as it is presented in \cite{tasFunctionalSystemArchitectures2016,ulbrichGraphbasedContextRepresentation2014}
architectures. Occlusions in specific situations need to be addressed
and may need objects representation. Several decision-making approaches
enable to take into account occluded areas. POMDP are used for handling
occlusions for static obstacles \cite{boutonScalableDecisionMaking2018}
as well as dynamic ones \cite{hubmannPOMDPManeuverPlanner2019,schornerPredictiveTrajectoryPlanning2019}.
The authors in \cite{hoermannEnteringCrossroadsBlind2017} uses several
layers of information: a dynamic grid, an object list and a map of
unobservable regions to assess likely collisions. Some articles handle
directly polygons of occluded areas \cite{narksriCrossingBlindIntersections2019},
in some cases the centerline is used \cite{hubmannPOMDPManeuverPlanner2019}
and in other cases the segment border is used \cite{orzechowskiTacklingOcclusionsLimited2018}.

When hidden areas appear, one needs to assess whether all of them
are relevant for a safe decision making. The notion of areas of interest
will be presented as there is a need to identify where relevant information
is missing. As demonstrated in \cite{sunBehaviorPlanningAutonomous2019},
the term ``Social perception'' is defined in order to infer information
in occluded zones from the behavior of road users. In specific situations,
additional information is inferred. A method patented \cite{determinevisibilitypatent}
shows the mechanism to determine a visibility distance by intersecting
the path of a vehicle from a map with the dynamic field of view. This
example shows a characterization of the following path of the vehicle
by two states: the free space that is of interest and the hidden one.
The Responsibility Sensitive Safety (RSS) model presented in \cite{rssmodel},
also highlights the spatial safety requirement with objects representation.
The concept of a safety distance is introduced for several situations.

Several works need a spatial representation handling occlusions and
objects for tasks like decision-making \cite{schornerPredictiveTrajectoryPlanning2019},
risk analysis \cite{yuOcclusionAwareRiskAssessment2019} or safety
evaluation \cite{hoermannEnteringCrossroadsBlind2017}. In \cite{orzechowskiTacklingOcclusionsLimited2018},
reachable sets are used for predicting the path and intention of a
potential vehicle that could be hidden in an occluded area. Particle
filters are also used in order to determine the plausibility of a
hidden vehicle \cite{narksriCrossingBlindIntersections2019}. Then,
as presented in \cite{hubmannGenericDrivingStrategy2016}, a global
method for motion planning could use non-misleading information provided
by such a context modeling module.

In this paper, the different levels of abstraction of the information
allows to identify the relevant parts of the situation that the autonomous
vehicle encounters. Then an analysis of the areas of interest is presented.
Additional information that can be inferred from occluded zones and
from objects are integrated in a unified representation.

\section{Interacting lane grid\label{sec:Approach}}

The three levels of abstraction provide a hierarchy of levels in which
several processes take place. Each level has a specific type of task
to achieve in a world model module.

\subsection{Strategic level of abstraction (high level)}

This level has the duty to compute and infer high-level information.
At this level the mission planning process provides the intended path
of the autonomous vehicle on a topological representation of the map.
From this intended path, one can define all the driving lanes that
may interact with the car during its cruise. Two types of interacting
lanes are differentiated. The primary order lanes are directly in
interaction,\textit{ i.e.}, shares a common driving space, with the
ego vehicle intended path. These lanes are the ones for which the
vehicle needs to have information, typically whether they are occupied
by another road user, for decision-making. The second order encompasses
lanes that have an indirect interaction with the autonomous vehicle,
in the sense that they have a direct interaction (\textit{i.e.}, primary
order) with the primary order lanes of the ego vehicle itself. The
vehicle does not necessarily need information about the second order
lanes to drive, but they are able to provide additional information
about the primary ones.

In this paper, we consider fours types of interactions that widely
cover common driving situations: lane keeping, changing, merging and
crossing interactions. In \tabref{Paths-of-interest} the interacting
paths are defined based on the topological layer of the road. In lane
keeping, the followed lane belongs to the primary order area of interest.
The changing lanes belong to the secondary order, as they are the
primary order lanes of the lane changing interaction type.

A similar approach can also be considered towards vulnerable road
users. A map with the cycle paths, sidewalks or pedestrian crossing
would enable to define additional types of interaction. These cases
are out of the scope of this work.

\subsection{Tactical level of abstraction (intermediate level) }

At the tactical level, the autonomous vehicle has to convert the topological
information provided at the strategic level into a metrical representation.
Moreover, the vehicle has to focus on areas of interest that are in
its vicinity in terms of time and distance.

Using a metrical HD map, the spatial areas of a lane can be extracted
as a 2D polyline, \textit{i.e.}, a sequence of line segments, representing
the center of the lane along with a width. The advantage of coding
the center of a lane compared to coding only the lane borders, \textit{e.g.},
with Lanelets~\cite{poggenhansLanelet2HighdefinitionMap2018}, is
that it is straightforward to discretize in the longitudinal direction.
By defining a threshold in terms of interacting distance, we prune
geometrically the interacting lanes and construct a spatial Area Of
Interest ($AOI$). The $AOI$ is therefore composed of the primary
$AOI^{(1)}$ and the secondary one $AOI^{(2)}$, all the space outside
of the $AOI=AOI^{(1)}\cup AOI^{(2)}$, denoted as $\overline{AOI}$,
is considered out of interest as there will be no interaction with
the ego vehicle.

The next step is to characterize the $AOI$ using perception information
from exteroceptive sensors such as cameras or LiDAR.The sensor setup
of the vehicle can be characterized beforehand in order to determine
its physical perception range. Prior to the perception itself, it
is possible to define the spatial Field Of View ($FOV$) of the vehicle.
Its complementary $\overline{FOV}$, corresponds to regions where
the vehicle has no means to perceive anything using its own embedded
sensors.

For decision-making at the tactical level, we propose to use a lane
level grid representation that discretizes the lanes within $AOI$
along their longitudinal direction.

\subsection{Operational level of abstraction (low level)\label{subsec:Operational-level}}

The lowest level of abstraction corresponds to a representation at
the sensor level. The space characterization is classically done using
grids~\cite{hoermannEnteringCrossroadsBlind2017} or parametric free
space~\cite{schreierCompactRepresentationDynamic2016}. The frame
of discernment $\Omega$, \textit{i.e.}, the space state, is then
decomposed into three categories: free ($F$), occupied ($O$) and
unknown ($U$).

\begin{figure}
\hspace*{\fill}\subfloat[\label{fig:charac_a}]{\includegraphics[width=0.4\columnwidth]{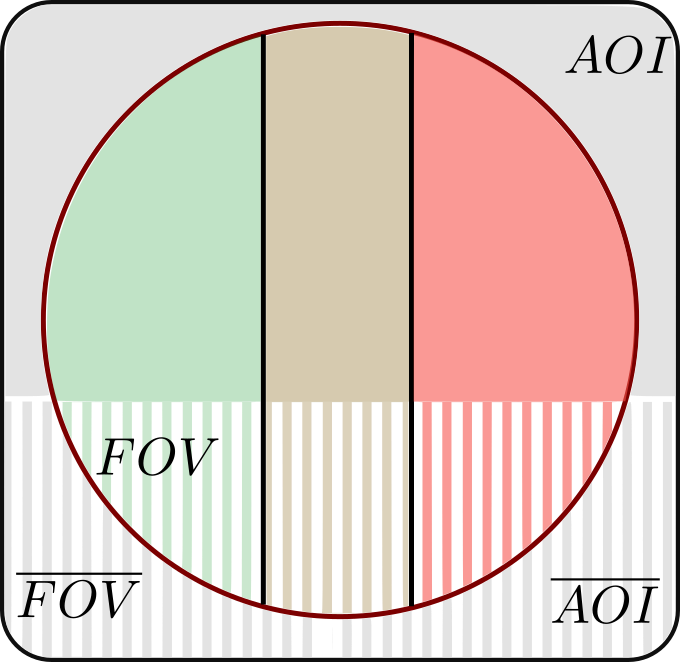}}\hfill{}\subfloat[\label{fig:charac_b}]{\includegraphics[width=0.4\columnwidth]{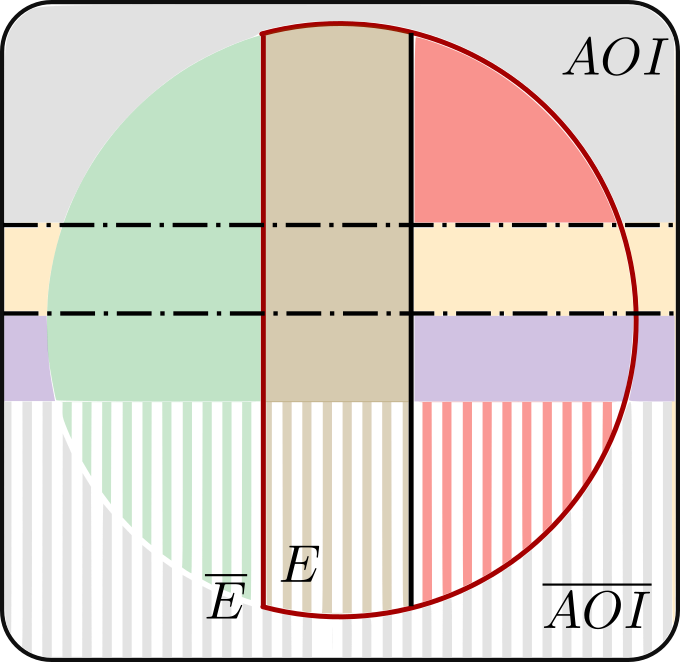}}\hspace*{\fill}\caption{Set diagram of an area of interest. Colors show the characterization
of the area of interest from \subsecref{Operational-level} for \figref{charac_a}
and from section \ref{sec:Interaction-objects} for \figref{charac_b}.\label{fig:Set's-diagram}}
\end{figure}

So far, the frame $\Omega$ has been decomposed in three different
ways: 
\begin{equation}
\Omega=AOI\cup\overline{AOI}=FOV\cup\overline{FOV}=F\cup O\cup U
\end{equation}
By crossing these three breakdowns, we obtain the space depicted in
\figref{charac_a}. One can see that the free and occupied sets are
necessarily inside the $FOV$. Conversely, the unknown set $U=\overline{F\cup O}$
has non-empty intersection with the $FOV$. Therefore, we will distinguish
the hidden set $H=U\cap FOV$, that represents the space within the
$FOV$ of the sensors but cannot be characterized because of occlusions,
from the rest of unknown space $U\cap\overline{FOV}$ that are out
of reach from the sensors range. As we are only interested in the
space inside the $AOI$, the remaining sets of interests becomes:
\begin{eqnarray}
F^{*} & = & F\cap AOI\\
O^{*} & = & O\cap AOI\\
H^{*} & = & H\cap AOI=U\cap FOV\cap AOI\\
U^{*} & = & U\cap\overline{FOV}\cap AOI
\end{eqnarray}
These four sets $F$, $O$, $H$, $U\cap\overline{FOV}$ are represented
in \figref{charac_a} by the colors green, brown, red and gray, respectively.
The subsets of interest $\left[\,\right]^{*}$are represented in filled
color whereas their complement is hatched.

\section{Interacting objects\label{sec:Interaction-objects}}

Static objects have an impact on the decision-making process. They
are typically well represented in occupancy grids. At the opposite,
dynamic objects\textit{ (e.g.}, other road users) interact with the
surrounding environment. More complex information can be inferred
from their presence. The detected objects that are outside of the
$AOI$ are discarded. On \figref{charac_b}, an entity defined as
a dynamic object is denoted $E$ and gives information about the occupied
$O$ and hidden $H$ space states. Its complementary $\overline{E}$
is all but dynamic objects.

\subsection{Safety area\label{subsec:Safety-area}}

When driving, the space in front of the vehicle needs to be obstacle-free
for obvious safety reason. This space is defined by a safety distance
that is an increasing function of the vehicle velocity. The faster
a vehicle is driving the more space needs to be kept as free in front
of it. This safety distance encompasses the driver's reaction time,
the minimum braking distance and safety regulations.

Therefore, when a vehicle is detected and its speed estimated, a safety
area at its front denoted $S$ is defined. In the case where this
safety area is visible and characterized as free (as it should be)
or occupied (it means that the theoretical safety distance is not
respected), then the safety property can be ignored as it is less
informative than the free or occupied knowledge. The safety property
is therefore only useful to further characterize an unknown area which
could be either hidden or out of the $FOV$. Its corresponding set
of interest is:
\begin{equation}
S^{*}=S\cap AOI
\end{equation}

This set is represented in yellow in \figref{charac_b}. 

\subsection{Protected area}

One important aspect of the interaction between road users is that
they impose physical constraints among each other. Once a vehicle
occupies a given space of the road, no other road users can cross
this space without causing an accident.

In our case study, we propose to use objects belonging to the second
order $AOI^{(2)}$ to further characterize the first order $AOI^{(1)}$.
Because the lanes in $AOI^{(2)}$ have a direct interaction, \textit{e.g.},
crossing, with the ones in $AOI^{(1)}$, a vehicle in $AOI^{(2)}$
may obstruct the circulation in $AOI^{(1)}$. This will happen when
a vehicle belongs to $AOI^{(2)}$ and occupies a spatial space that
also belongs to a lane in $AOI^{(1)}$. All the space behind this
vehicle along the lane in $AOI^{(1)}$is protected in the sense that
no other road users can go through the obstacle. This protected area
is denoted $P$.

Like the safety case, the protected property is only relevant within
the unknown category and can be modeled by the following set of interest:
\begin{equation}
P^{*}=P\cap AOI
\end{equation}

This set is represented in purple in \figref{charac_b}.

\Tabref{Characterization-sets-table} depicts only the classification
of the $AOI$. As areas of interest are linked to the ego-vehicle
main path, their representation can also be seen as a graph of successive
states whose edges are represented by a distance to the ego-vehicle
state. A whole cell can be classified as been occupied regardless
of the object size. Indeed, the distance to this occupied state and
the area of interest which it belongs will be supplied to the decision-making.

This characterization process brings information and enables to structure
the environment understanding. It should be noticed that if additional
information characterization can be refined, the model is able to
extend the process to integrate potentially new sets of information.

For clarity, in the rest of the paper, everything outside of the $AOI$
(dashed set in \figref{Set's-diagram}) will be ignored and the notation
$\left[\,\right]^{*}$ will be omitted.

\begin{table}
\begin{centering}
\begin{tabular}{c|c|c|c|c|c|c|}
\cline{2-7} \cline{3-7} \cline{4-7} \cline{5-7} \cline{6-7} \cline{7-7} 
 & \multicolumn{4}{c|}{{\scriptsize{}${\normalcolor FOV}$}} & \multicolumn{2}{c|}{{\scriptsize{}$\overline{FOV}$}}\tabularnewline
\cline{2-5} \cline{3-5} \cline{4-5} \cline{5-5} 
 & \multirow{2}{*}{{\scriptsize{}Free}} & \multirow{2}{*}{{\scriptsize{}Occupied}} & \multicolumn{2}{c|}{{\scriptsize{}~~~~Hidden~~~~}} & \multicolumn{2}{c|}{}\tabularnewline
\cline{4-7} \cline{5-7} \cline{6-7} \cline{7-7} 
 &  &  & \multicolumn{4}{c|}{{\scriptsize{}Unknown}}\tabularnewline
\hline 
\multicolumn{1}{|c|}{{\scriptsize{}${\normalcolor AOI}$}} & {\scriptsize{}\cellcolor{green}$F^{*}$} & {\scriptsize{}\cellcolor{brown}$O^{*}$} & {\scriptsize{}\cellcolor{red}$H^{*}$} & \multicolumn{2}{c|}{{\scriptsize{}\cellcolor{yellow}$S^{*}$: Safety}} & {\scriptsize{}\cellcolor{grey}$U^{*}$}\tabularnewline
\cline{5-6} \cline{6-6} 
\multicolumn{1}{|c|}{} & {\scriptsize{}\cellcolor{green}} & {\scriptsize{}\cellcolor{brown}} & {\scriptsize{}\cellcolor{red}} & \multicolumn{2}{c|}{{\scriptsize{}\cellcolor{purple}$P^{*}$: Protected}} & {\scriptsize{}\cellcolor{grey}}\tabularnewline
\hline 
\end{tabular}
\par\end{centering}
\caption{Characterization sets table of areas of interest\label{tab:Characterization-sets-table}}
\end{table}

\section{Case studies}

To illustrate the concepts introduced in the previous sections, let
us consider the two case studies pictured in \figref{use_case} and
\figref{overtaking}. The ego vehicle is represented by the blue car
which follows the dotted intend path. The set of all colored lanes
constitutes the $AOI$ as defined in \tabref{Paths-of-interest},
these lanes are discretized in cells along their longitudinal direction.
The light blue circle depicts the $FOV$ of the ego vehicle. The orange
cars are other road users.

\subsection{T-intersection}

\begin{figure}
\includegraphics[width=0.9\columnwidth]{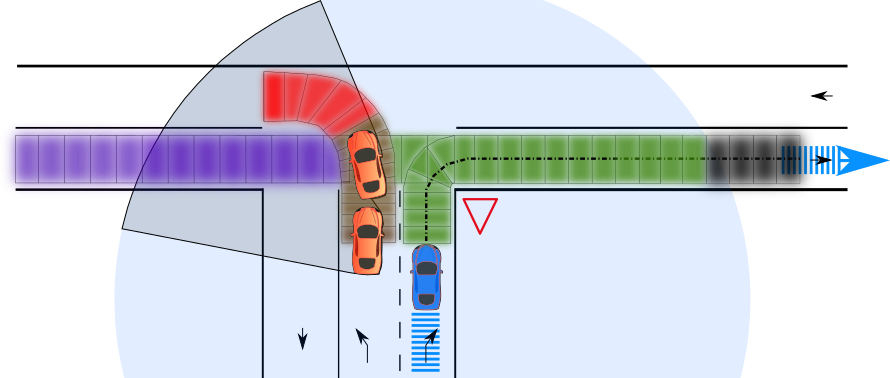}

\caption{T-intersection. Areas of interest are modeled and characterized: Free
(green), Occupied (brown), Out of range (black), Hidden (red), Protected
(purple) \label{fig:use_case}}
\end{figure}

The first situation is at a T-intersection as shown in \figref{use_case}.
The ego-vehicle intends to turn right and has to give way to the vehicles
coming from the left merging lane. The two red cars on the adjacent
left lane do not interact with the ego vehicle but cross the merging
lane and cause occlusions. This example represents a situation where
the ego-vehicle should be able to take a decision even under occlusion
at a give-way intersection.

In this situation, the ego-vehicle has two primary order interacting
lanes: a lane keeping one, \textit{i.e.}, the right-turn lane, and
a lane merging one, \textit{i.e.}, the lane coming from the left.
The adjacent left lane of the ego vehicle is a secondary order interacting
lane as it crosses one of the primary ones. The set of all these lanes
are then pruned down to a distance horizon of interest to constitute
the $AOI$. One can note that the second order lane is pruned very
early as it becomes out of interest. Within the visible area the free
cells ($F$) are shown in green. The cells occupied by the red cars
are in brown ($O$). The black cells are unknown ($U$) as they are
out of the $FOV$. The red cells are also unknown but, as they are
inside the $FOV$ and not observed because of occlusions caused by
obstacles, they are categorized as hidden ($H$). Last, the purple
cells highlight the protected ($P$) area. Like $U$ and $H$, $P$
is not observed, but its space is physically obstructed by the presence
of other road users. It should be noted that the protected area may
reach outside of the $FOV$.

\subsection{Overtaking}

\begin{figure}
\hfill{}\subfloat[\label{fig:overtaking_a}]{\includegraphics[width=0.4\columnwidth]{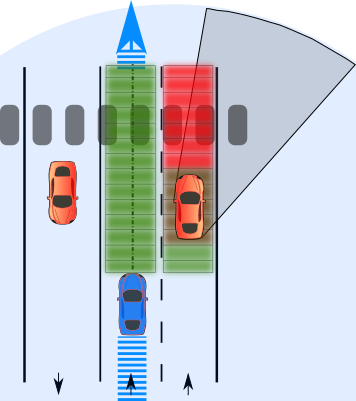}}\hfill{}\subfloat[\label{fig:overtaking_b}]{\includegraphics[width=0.4\columnwidth]{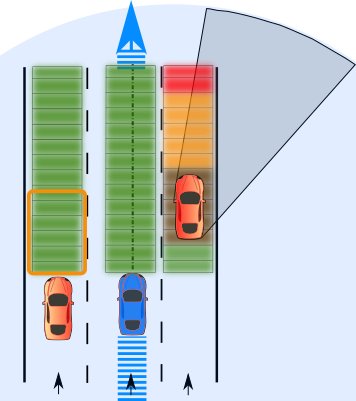}

}\hfill{}\hfill{}\caption{Straight road with only lane change situations. Areas of interest
are modeled and characterized : Free (green), Occupied (brown), Hidden
(red), Safety zone (orange). On \figref{overtaking_a}, the right
side vehicle is stopped . On \figref{overtaking_b} the right side
vehicle has a constant speed.\label{fig:overtaking}}
\end{figure}

Let us consider a second situation where the ego vehicle intends to
overtake a vehicle positioned on its right side as pictured in \figref{overtaking}.
As explained in section \ref{subsec:Safety-area}, a safety area is
defined in front of every detected object by using its estimated speed.
It is assumed here that vehicles constantly keep safe inter-distance
with other vehicles so they can safely perform emergency braking to
avoid collision in case of dangerous situations. It means that it
is likely that the immediate space in front of any vehicle is free.
In \figref{overtaking_a} the orange vehicle on the right is stopped.
In this case, there is no safety space, therefore the unknown area
in front of it is characterized as hidden (red). In \figref{overtaking_b},
the same vehicle is moving, therefore a portion of the space in front
of it is safe (orange). Note that for the car on the left side of
the ego vehicle, the safe property of the cells in front of it (orange
rectangle) is ignored as the cells have already been classified as
free, which is more informative.

\subsection{Experimental setup and implementation}

The methodology presented in the previous sections has been implemented
on a Renault ZOE experimental vehicle of the Heudiasyc laboratory.
Experiments have been conducted in the city of Compi�gne, France,
where an HD map has been constructed. The lanes in the map are represented
by their center as a polyline with their nodes being geo-referenced
with centimeter accuracy. The borders of each lane are also encoded.
The graph structure behind the map is used at both the topological
and metrical levels. It enables to compute the ego vehicle intended
path as well as all the primary and secondary order interacting lanes.
For the computation of the $AOI$, we set a longitudinal distance
of interest of $100$m and an arbitrary cell discretization step of
$1$m.

A NovAtel SPAN-CPT solution, combining an IMU and GNSS with RTK corrections,
was used to have an accurate localization of the ego-vehicle. A Velodyne
VLP32-C LiDAR was used for the environment perception. This sensor
has a $360^{\circ}$ field of view with a theoretical range of $100$
meters. A simple geometric ground fitting and clustering algorithm
was used to measure the free space and object surrounding the ego
vehicle \cite{groundfitting}. For the computation of the safety space,
an emergency braking model with a deceleration of $a=-6\,\text{m \ensuremath{\cdot\text{s}^{-2}}}$
is used. The safety distance associated to a vehicle driving at a
speed $V$ is computed as $d_{\text{safe}}=-V^{2}/(2a).$

The ego-vehicle and paths of interest are extracted recursively from
the HD map as they can be preprocessed thanks to the topological layer.
Areas of interest are represented by polygons using the center line
with its width. The ROS middle-ware was used for the implementation
and the Shapely Python library was used for the geometric operations
between areas of interest and the free space polygons. The difference
characterizes the hidden space. \Figref{Rviz-display} shows the results
from real data in a round about scenario.

\vspace{-1pt}

\subsection{Discussions}

\begin{figure}
\begin{centering}
\begin{tabular}{cc}
\includegraphics[width=0.45\columnwidth]{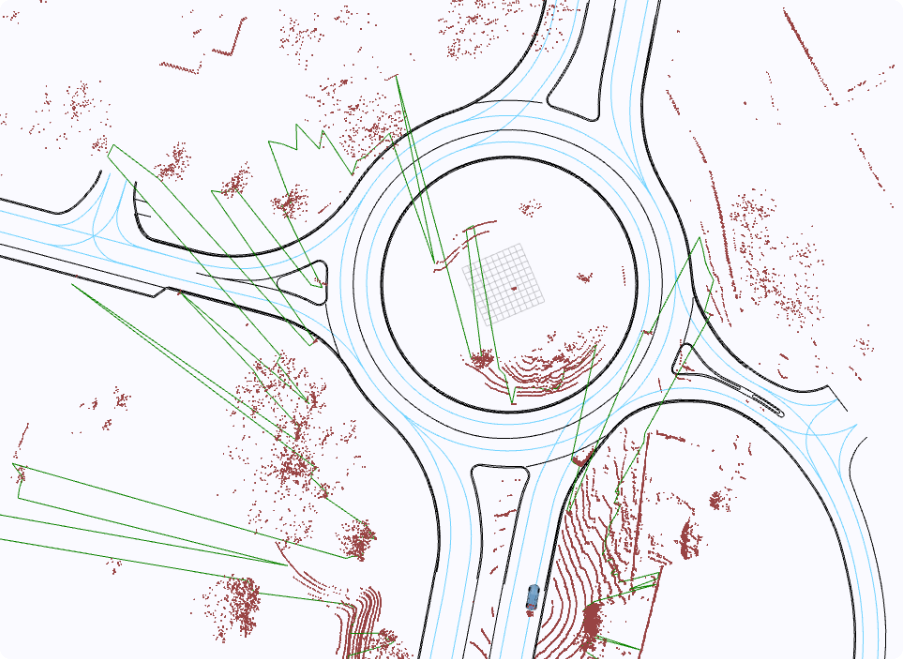}\label{fig:top_left} & \includegraphics[width=0.45\columnwidth]{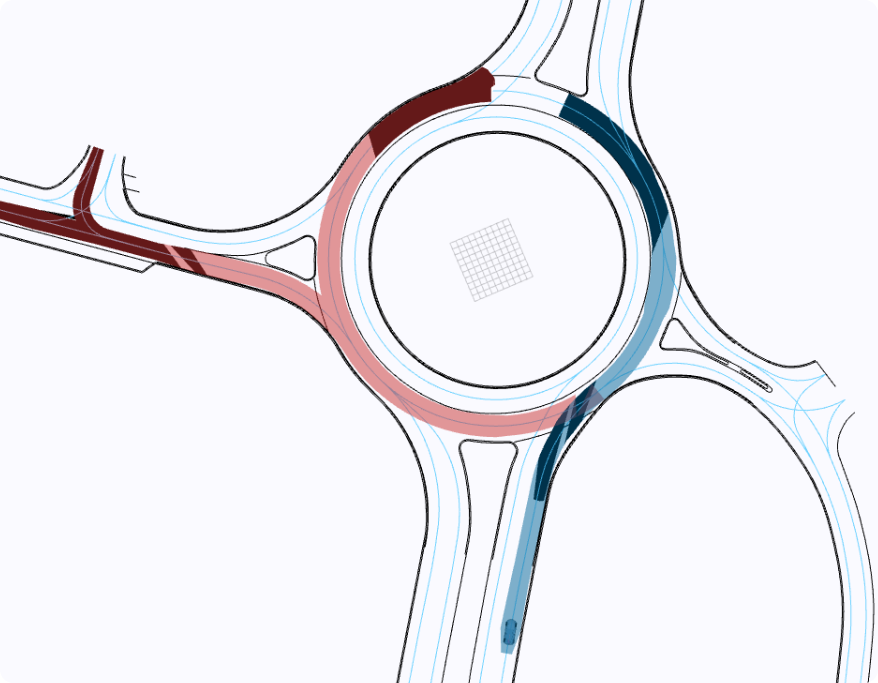}\label{top_right}\tabularnewline
\includegraphics[width=0.45\columnwidth]{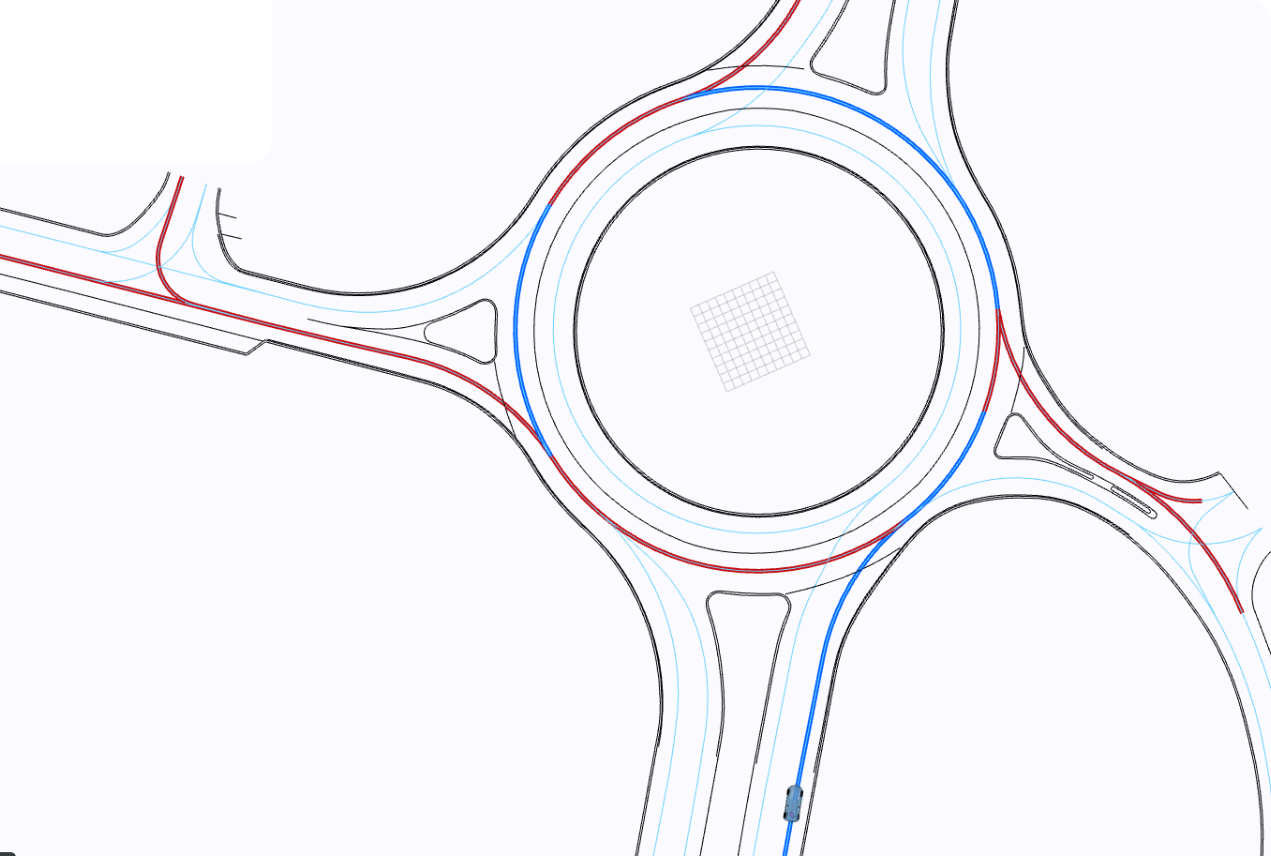}\label{bot_left} & \includegraphics[width=0.45\columnwidth]{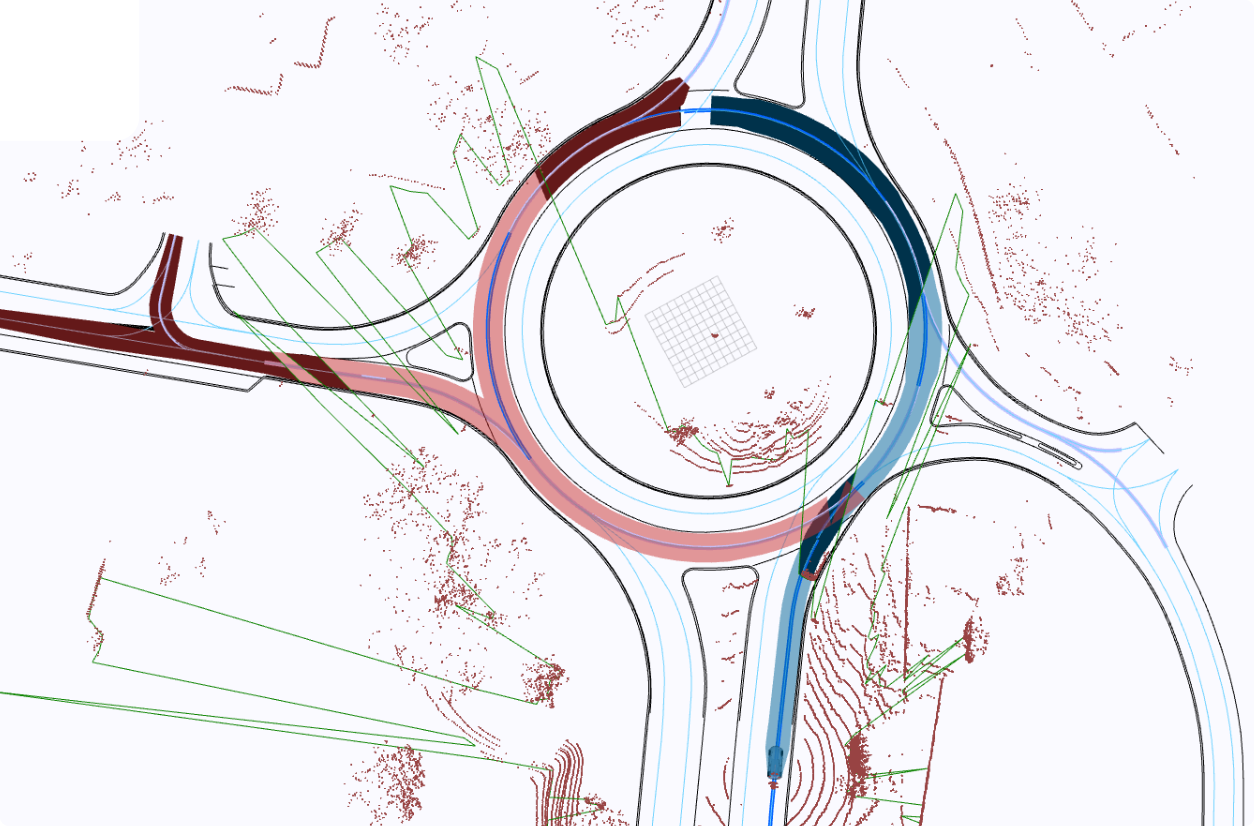}\label{bot_right}\tabularnewline
\end{tabular}
\par\end{centering}
\caption{Rviz display of real data using ROS. The ego-vehicle is displayed
in blue. The figure on the top left shows the free-space polygon (Green),
operational level. The figure on the top right illustrates the $AOI$
(blue : lane keeping, red: lane merging), tactical level. The figure
on the bottom left depicts the path (as a graph) of the ego-vehicle
(blue) and its primary order interacting paths (red). The figure on
the bottom right shows final representation. In this scenario, only
the hidden space is shown characterized (black) thanks to the free-space
polygon.\label{fig:Rviz-display}}
\end{figure}

\Figref{Rviz-display} shows the result from real data in a roundabout
scenario at a given time.

The hypothesis made in our experiments can easily be extended to more
complex assumptions. The interaction distance of the $AOI$ can be
set dynamically depending on the ego vehicle dynamic state. The $AOI$
cell discretization step can also be set dynamically instead of a
constant. The further we are, the more the step increases. The free
space can be replaced by a more reliable occupancy grid that integrates
information over time and more sophisticated perception algorithms,
\textit{e.g.}, deep learning-based, can be used of object detection
and tracking. Extension to include vulnerable road users can also
be done based on the same principles.

It is important to note that the illustrations shown for the two case
studies are instantaneous representation of the world at a given time.
For the T-intersection case, the decision of the ego vehicle to enter
in the intersection relies on the fact that the protected area remains
protected during the whole duration of the insertion maneuver. Our
representation space should therefore also be used for prediction
purposes where the predicted positions of the tracked objects are
used instead.
\vspace{-10pt}

\section{Conclusion}

In this work, an intermediate lane level information representation
approach has been proposed with an interacting lane grid. Interacting
lanes are extracted from the map and define the interactions encountered
by the autonomous vehicle. These interacting areas enable the autonomous
vehicle to focus on relevant space parts and to look for the corresponding
potential interactions. This lane level grid is discretized and each
cell is classified from most informative state: free, occupied, protected,
safety zone, to less informative: hidden, out of field of view. This
process for the cell grid classification has been depicted. We build
information on space representation from spatial perception (as grids)
and from objects. This information representation is provided for
decision-making at the maneuver planner level.

In future work, objects detection, tracking and behavior prediction
will be implemented in order to improve cells classification and situation
predictions will be explored.

\vspace{-10pt}

\bibliographystyle{IEEEtran}
\bibliography{biblio}

\end{document}